\documentclass{article}

\usepackage[final]{nips_2017b}

\usepackage[utf8]{inputenc} 
\usepackage[T1]{fontenc}    
\usepackage{hyperref}       
\usepackage{url}            
\usepackage{booktabs}       
\usepackage{amsfonts}       
\usepackage{nicefrac}       
\usepackage{microtype}      
\usepackage{natbib}
\usepackage{graphicx}
\usepackage{amsopn}
\usepackage{amsmath}

\usepackage[ruled,vlined]{algorithm2e}

\title{Curiosity-driven reinforcement learning with homeostatic regulation}

\author{
  Ildefons Magrans de Abril\\
  ARAYA, Inc.\\
  Tokyo\\
  \texttt{ildefons@araya.org} \\
  \And
  Ryota Kanai \\
  ARAYA, Inc.\\
  Tokyo\\
  \texttt{kanair@araya.org} \\
}

\begin{document}

\maketitle

\begin{abstract}
We propose a curiosity reward based on information theory principles and consistent with the animal instinct to maintain certain  critical  parameters  within  a  bounded  range. Our experimental validation shows the added value of the additional homeostatic drive to enhance the overall information gain of a reinforcement learning agent interacting with a complex environment using continuous actions. Our method builds upon two ideas: i) To take advantage of a new Bellman-like equation of information gain and ii) to simplify the computation of the local rewards by avoiding the approximation of complex distributions over continuous states and actions.
\end{abstract}

\section{Introduction}
\label{introduction}

Within a reinforcement learning setting \citep{sutton1998reinforcement}, a reward signal indicates a particular momentary positive experience and it serves to constrain the long-term agent behavior. Extrinsic rewards are generated by an external oracle and they indicate how well the agent is interacting with the environment (e.g. videogame score, portfolio return). On the other hand, intrinsic rewards are generated by the agent itself and they indicate a particular internal event sometimes implemented as a metaphor of an animal internal drive \citep{chentanez2005intrinsically}.

There are many intrinsic rewards and most of them can be characterized by how they affect the information flow between the environment and the agent. In one side of the spectrum, information is pushed from the agent to the environment, for instance, by rewarding actions that lead to predictable consecutive sensor readings \citep{montufar2016information} or by rewarding reaching states from where the agent actions have a large influence in determining the future state (i.e. empowerment \citep{jung2011empowerment,mohamed2015variational}). 

On the other side, information is encouraged to efficiently move from the environment to the agent. These rewards motivate the agent to explore its environment by taking actions leading to an improvement of its internal models. \cite{jurgen1991possibility} proposed an online learning agent equipped with a curiosity unit measuring the Euclidian distance between the observed state and the model prediction. Recently, \cite{pathak2017curiosity} extended the curiosity functionality to accommodate agents with high dimensional sensory inputs by adding a representation network able to filter out information from the observed state that is not relevant to predict how the agent actions affect the future state. \citep{houthooft2016vime} presented an exploration reward bonus based on information gain maximization computed using a variational approximation of a Bayesian neural network. \cite{lopes2012exploration} discussed an exploration reward bonus that encourages the learning progress over the last few experiences instead of the immediate agent surprise. \cite{bellemare2016unifying} differ in the sense that the agent is not learning a forward model but a probability density function about the states visited by the agent together with a lower bound on  the information gain associated with the agent exploratory behavior. 





A common denominator of all methods is an intrinsic reward function that encourages actions with a high information gain potential. This behavior is consistent with the innate drive to explore of humans and other animals. However, it is probably incomplete because it is not able to accommodate the also innate animal desire to maintain certain critical parameters within a  bounded range (i.e. physiological constants) \citep{carver1998self}. To achieve that goal in an uncertain environment, the animal has to trade its curiosity drive with the need to act according to familiar patterns that guarantee the required stimuli (e.g. food, water, heat,...). 

In the following sections we derive our approach from information theory considerations. It extends the typical heterostatic curiosity reward with an additional term. Our novel extension simulates an animal homeostatic drive to keep a "familiar" behavior. More precisely, we validate the value of regulating a purely heterostatic curiosity reward with an homeostatic drive in the context of an agent trying to learn how its environment responds to its actions. The concept of homeostatic regulation in social robots was first proposed in \citep{breazeal2004designing}. 

\section{Background}
\label{background}

This paper assumes a typical reinforcement learning setup where an agent interacts with the environment at discrete time steps, it observes an state $S_t$ $\in$ $S$ and it acts on the environment with action $A_t$ $\in$ $A$ according to a control policy $\pi(A_t|S_t)$, the main goal of an information gain agent is to learn a forward model that explains how the environment reacts to its actions. 

Recently, \cite{tiomkin2017unified} presented a recursive expression to describe the information transferred from a sequence of states to the following sequence of actions when an agent interacts open-endedly with a Markovian environment (i.e.  transition probability function $\sim$ $P(S_{t+1}|S_t,A_t)$). Figure \ref{fig:diagram} shows a conceptual diagram of the information gain process and equation \ref{eqn:infogain} presents the recursive expression for information gain:

\begin{figure}[h!]
  \centering
  \includegraphics[width=0.30\textwidth]{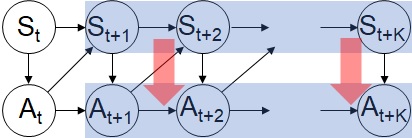}
  {\caption{Conceptual diagram of the information gain process: dark thin arrows are causal dependencies and large arrows show how information flows from a sequence of observed states to the corresponding future sequence of agent actions.}
  \label{fig:diagram}}
\end{figure}

\begin{eqnarray}
  \label{eqn:infogain}
   && InformationGain(s_t)=I(S_{t+1:t+K} \rightarrow A_{t+1:t+K} || S_{t:t+K-1},A_t) \nonumber \\
  &&=I(S_{t+1};A_{t+1}|S_{t},A{t}) \nonumber \\
  &&+<I(S_{t+2:t+K} \rightarrow A_{t+2:t+K} || S_{t+1:t+K-1},A_{t+1})>_{P(S_{t+1},A_{t+1}|S_t,A_t)} ,
\end{eqnarray}

where $I(A \rightarrow B || C)$ is the causally conditioned directed mutual information \citep{kramer1998directed} and $A$/$S_{t+1:t+K}$ is the sequence of actions/states of length $K$ starting at time $t+1$.

It is important for our approach that the definition of information gain can be expressed recursively with a similar decomposition as the Bellman equation, where $I(S_{t+1};A_{t+1}|S_{t},A{t})$ would be the agent reward obtained at time t by taking action $A_t$ at state $S_t$ and the second component would be the average discounted reward (with discount factor 1). Computing the conditional mutual information of the local reward requires the approximation of the corresponding probability distributions \citep{mohamed2015variational,tiomkin2017unified}. When actions and/or states are discrete, we can approximate them for instance using a neural network with a softmax output layer. However, it’s much harder when states and actions are continuous, especially when the state space is very high dimensional (e.g. video stream). 

We propose a more practical method to implement information gain in a reinforcement learning setting when states and actions are continuous and we show how our approach, derived from information theory principles, extends the narrow view of existing approaches by compensating the heterostacity drive encouraged by the curiosity reward with an additional homeostatic drive.

\section{Approach}
\label{approach}

Our method builds upon two ideas: i) To take advantage of the Bellman like equation of information gain (equation \ref{eqn:infogain}) to justify the use of a reinforcement learning algorithm as underlying mechanism to explore the environment; ii) To simplify the computation of the local reward defined as $I(S_{t+1};A_{t+1}|S_{t},A{t})$ by avoiding the approximation of complex distributions over continuous states and actions.

We implemented the first idea using a state of the art RL algorithm that works well with continuous actions. We chose the Deep Deterministic Policy Gradient algorithm \citep{lillicrap2015continuous} but other options are also feasible. This algorithm finds a deterministic control policy that maximize the expected sum of discounted rewards. When $\gamma = 1$, episode length is $K$ and reward function is $I(S_{t+1};A_{t+1}|S_{t},A{t})$, then our agent explores the environment by maximizing the information gain as expressed in equation \ref{eqn:infogain}.

\begin{figure}[h!]
  \centering
  \includegraphics[width=0.50\textwidth]{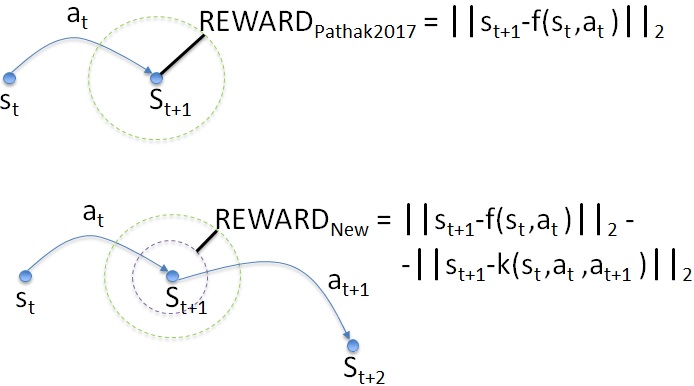}
  {\caption{Semantic of the curiosity reward with homeostatic regulation and comparisson with respect to a state of the art curiosity reward based on the Euclidian distance between the observed state and the model prediction (E.g. \citep{jurgen1991possibility,pathak2017curiosity}).}
  \label{fig:diagram2}}
\end{figure}

We can express this reward as the reduction of entropy in the future state $S_{t+1}$. Then, because we are able to know exactly the current state and due to the deterministic nature of the control policy inferred by the DDPG algorithm, we use the concrete state $s_t$ and actions $a_t$ and $a_{t+1}$ instead of the random variables $S_t$, $A_t$ and $A_{t+1}$ respectively to compute the reward. Finally, we approximate the reduction of entropy in the future state $S_{t+1}$ as the reduction of the prediction error in the future state. Equation \ref{eqn:simplification} formalizes this approximation:

\begin{eqnarray}
  \label{eqn:simplification}
   && I(S_{t+1};A_{t+1}|s_{t},a_{t})= H(S_{t+1}|S_{t},A_{t}) - H(S_{t+1}|S_{t},A_{t},A_{t+1}) \nonumber \\
   && \approx H(S_{t+1}|s_{t},a_{t}) - H(S_{t+1}|s_{t},a_{t},a_{t+1}) \nonumber \\
  && \approx ||s_{t+1} -\hat{s}_f||_2-||s_{t+1} -\hat{s}_k||_2
\end{eqnarray} 

where $\hat{s}_f = f(s_t,\pi(s_{t})=a_t)$ and $\hat{s}_k = k(s_t,\pi(s_{t})=a_t,\pi(s_{t+1})=a_{t+1})$ are the future state predictions by the forward and extended forward models respectively. The extended forward model takes advantage of knowing the action that the agent will take in the future state to improve the prediction about this future state. This approximation captures the relevant semantic with much lower computational cost. Interestingly, the internal models $f(.)$ and $k(.)$ can be easily implemented with deep neural networks and suit well an agent with high-dimensional input streams. Figure \ref{fig:diagram2} is a graphical representation of the semantic of the new curiosity reward with homeostatic regulation and how it compares with respect to a state of the art curiosity reward based on the Euclidian distance between the observed state and the model prediction (E.g. \citep{jurgen1991possibility,pathak2017curiosity}).

Our new curiosity reward has two components: 1) Heterostatic motivation: similarly to an state of the art work based on the Euclidean distance, the first component of our reward encourages taking actions that lead to large forward model errors. This first component implements the heterostatic drive. In other words, the tendency to push away our agent from its habitual state; 2) Homeostatic motivation: the second component is our novel contribution. It encourages taking actions $a_t$ that lead to future states $s_{t+1}$ where the corresponding future action $a_{t+1}$ gives us additional information about $s_{t+1}$. This situation happens when the agent is "familiar" with the state-action pair: $\{s_{t+1},a_{t+1}\}$. Therefore, our new reward encourage the agent to move towards regions of the state-action space that simultaneously deliver large forward model errors and that are "known/familiar" to the agent. In other words it implies a priority sampling strategy towards "hard-to-learn" regions of the state-action space. 


We further generalize this reward by adding an hyper-parameter $\alpha>0$ that controls the importance of the of the homeostatic bonus. Finally, we should note that the reward function is non-stationary due to the continuous learning of $f$ and $k$. For that reason we $z$-normalize the reward using a mean and standard deviation computed at the end of each of episode using all available samples:

\begin{eqnarray}
  \label{eqn:finalreward1}
   && R(s_t)=  \frac{IG_{\alpha}(s_t)-\mu_{ig}}{\sigma_{ig}}
\nonumber \\
   && IG_{\alpha}(s_t)=||s_{t+1} -\hat{s}_f||_2-\alpha||s_{t+1} -\hat{s}_k||_2
\end{eqnarray} 

Algorithm \ref{alg:main} summarizes the overall logic of our curiosity agent. It follows an architecture similar to \cite{pathak2017curiosity}:

\begin{algorithm}
\SetAlgoLined
\KwResult{Forward model: $f(s,a)$ }
 Initialization of $f, k, DDPG$ parameters including $\pi$\;
 Initialization of random exploration probability $\epsilon$\;
 \For{episode i}{
 Initialize environment: initial state $s_0$ according to experiment strategy (see section \ref{results})\;
 \For{step t}{
 	Generate $a_t=\pi(s_t)$ (random according to $\epsilon$)\;
 	Sample $s_{t+1} \sim P(.|st, at)$\;
 	Get reward $r_t$ according to equation \ref{eqn:finalreward1}\;
 	Add $\{s_t,a_t,s_{t+1},r_t\} $ to replay buffer (RB)\;
 	Sample mini-batch $MB \sim RB$\;
 	Train $f, k$ and DDPG networks (including $\pi$);\
 	
  }
 }
 
\caption{Curiosity-driven reinforcement learning with homeostatic regulation}
\label{alg:main}
\end{algorithm}

\section{Results}
\label{results}

Our experimental validation presents two examples where both curiosity and homeostatic drives are superior to learn a forward model. Our validation hypothesis is that exploring an environment with several non-linearities could be optimized by regulating the agent curiosity with an homeostatic drive. More precisely, it could help by prioritizing the exploration of the state-action space according to how hard is to learn each region.    

\begin{figure}[h!]
  \centering
  \includegraphics[width=0.30\textwidth]{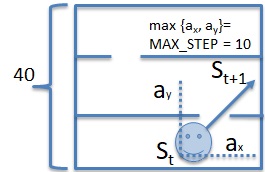}
  {\caption{Scheme of our 3 room environment.}
  \label{fig:diagram3}}
\end{figure}

To test our hypothesis, we use a 3 room environment (40x40), where an agent, able to sense its exact position, learns a control policy according to the DDPG algorithm with the reward presented in equation \ref{eqn:finalreward1} and a probability of taking a random action equal to $0.5$. The agent starts every episode in a random state and it runs for 10 steps (with max length step=10). We have implemented the forward model $f$ and the extended forward model $k$ as feed forward neural networks with 2 hidden layers with 64 hidden units each. We store the agent traces and we train the agent and the internal models at the same time following the same architecture of \citep{pathak2017curiosity}.  Figure \ref{fig:diagram3} shows an scheme of our environment. 

\begin{figure}[h!]
  \centering
  \includegraphics[width=0.90\textwidth]{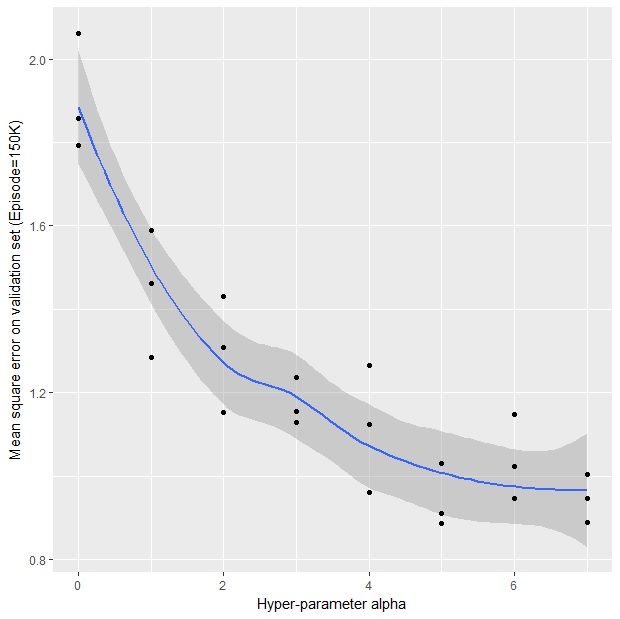}
  {\caption{Accuracy of the forward model learned by the agent as a function of $\alpha$ (measured according to the mean square error on the validation set).}
  \label{fig:resultsexperiment1}}
\end{figure}

\begin{figure}[h!]
  \centering
  \includegraphics[width=1.0\textwidth]{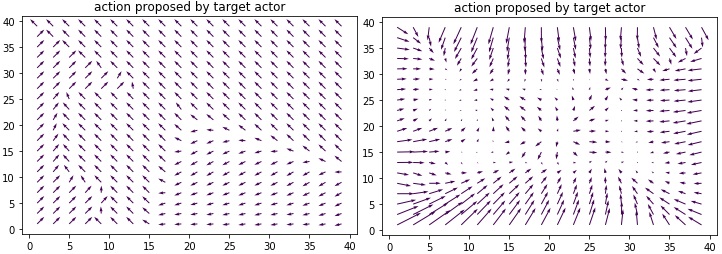}
  {\caption{Flow diagram of the control policy learned after 10K episodes with $\alpha=0$ (left) and $\alpha=7$ (right) respectively.}
  \label{fig:resultsexperiment1b}}
\end{figure}

In our first experiment we study the accuracy of the final forward model as a function of $\alpha$. We check the prediction accuracy using a pool of $10^7$ randomly generated samples. We run our agent using different values of $\alpha \in \{0,1,2,3,4,5,6,7\}$  for 150$K$ episodes and we do each experiment 3 times. Figure \ref{fig:resultsexperiment1} shows how we can improve the environment sampling efficiency by increasing the homeostatic component of the reward (i.e. $\alpha$). Figure \ref{fig:resultsexperiment1b} shows a diagram of the policy learned after 10K episodes with $\alpha=0$ and $\alpha=7$ respectively. We can clearly appreciate that, when $\alpha$ is large, the agent tends to position itself where there are a larger number of non-linearities (i.e. the "doors"). This agent behavior enhances the learning of complex regions by leveraging a more intense random exploration where it is most required. We should also discuss that for this particular experiment, a pure random sampling strategy achieves a mean square error on the validation set of 0.67 which is slightly better than the best result obtained with $\alpha=7$ (0.87). However, this is not a fair comparison because every episode starts in a different position which enables a pure random agent to reach every spot of the environment by simply random walking its local surroundings.

\begin{figure}[h!]
  \centering
  \includegraphics[width=0.90\textwidth]{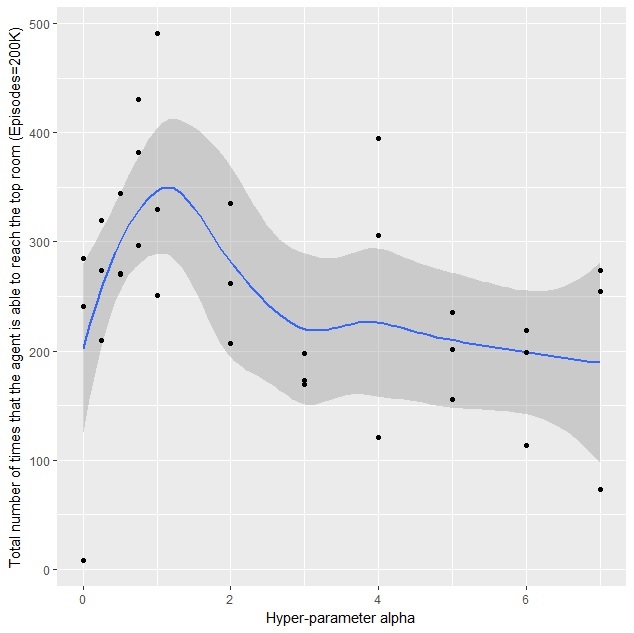}
  {\caption{Total number of times that the agent is able to reach the top room as a function of $\alpha$ when it starts every episode in a random position of the bottom room.}
  \label{fig:resultsexperiment2}}
\end{figure}

We performed a second experiment using the same environment described in Figure \ref{fig:diagram3}, but in this case the agent starts every episode in a random state of the bottom room. We want to understand whether the homeostatic reward is able to enhance the acquisition of innovative environment samples by counting how many times the agent is able to traverse 2 doors and reach the top room. Figure \ref{fig:resultsexperiment2} shows how we can improve the acquisition of challenging environment states by optimizing the contribution of the homeostatic reward component. In this case, a pure random sampling strategy running for 150$K$ episodes only reach the top room a total average of 145 times which is far below any other total average achieved with a non-random strategy with any $\alpha$.  




\section{Conclusions and future work}
\label{conclusions}

We presented an exploration approach that implements two apparently inconsistent animal drives: 1) the innate drive to explore (heterostatic behavior) and 2) the desire to maintain certain critical parameters stable. We derive an intrinsic reward function from information theory principles that generalize an state of the art method and we present experimental results to validate the superior exploration behavior of a join homeostatic and heterostatic drive with respect to a pure curiosity/heterostic approach. In future work, we want to explore meta-learning strategies to dynamically adjust the contribution of the homeostatic drive (i.e. $\alpha$) as well as the percentage of random exploration as a function of the agent progress in learning a the forward model.

\newpage

\subsubsection*{Acknowledgments}

This work was supported by JST CREST Grant Number JPMJCR15E2, Japan.

\bibliographystyle{plainnat}
\bibliography{IGIldefons}

\begin{thebibliography}{15}
\providecommand{\natexlab}[1]{#1}
\providecommand{\url}[1]{\texttt{#1}}
\expandafter\ifx\csname urlstyle\endcsname\relax
  \providecommand{\doi}[1]{doi: #1}\else
  \providecommand{\doi}{doi: \begingroup \urlstyle{rm}\Url}\fi

\bibitem[Bellemare et~al.(2016)Bellemare, Srinivasan, Ostrovski, Schaul,
  Saxton, and Munos]{bellemare2016unifying}
Marc Bellemare, Sriram Srinivasan, Georg Ostrovski, Tom Schaul, David Saxton,
  and Remi Munos.
\newblock Unifying count-based exploration and intrinsic motivation.
\newblock In \emph{Advances in Neural Information Processing Systems}, pages
  1471--1479, 2016.

\bibitem[Breazeal(2004)]{breazeal2004designing}
Cynthia~L Breazeal.
\newblock \emph{Designing sociable robots}.
\newblock MIT press, 2004.

\bibitem[Carver and Scheier(1998)]{carver1998self}
Charles~S Carver and Michael~F Scheier.
\newblock On the self-regulation of behavior. 10.1017.
\newblock \emph{CBO9781139174794}, 1998.

\bibitem[Chentanez et~al.(2005)Chentanez, Barto, and
  Singh]{chentanez2005intrinsically}
Nuttapong Chentanez, Andrew~G Barto, and Satinder~P Singh.
\newblock Intrinsically motivated reinforcement learning.
\newblock In \emph{Advances in neural information processing systems}, pages
  1281--1288, 2005.

\bibitem[Houthooft et~al.(2016)Houthooft, Chen, Duan, Schulman, De~Turck, and
  Abbeel]{houthooft2016vime}
Rein Houthooft, Xi~Chen, Yan Duan, John Schulman, Filip De~Turck, and Pieter
  Abbeel.
\newblock Vime: Variational information maximizing exploration.
\newblock In \emph{Advances in Neural Information Processing Systems}, pages
  1109--1117, 2016.

\bibitem[Jung et~al.(2011)Jung, Polani, and Stone]{jung2011empowerment}
Tobias Jung, Daniel Polani, and Peter Stone.
\newblock Empowerment for continuous agent—environment systems.
\newblock \emph{Adaptive Behavior}, 19\penalty0 (1):\penalty0 16--39, 2011.

\bibitem[Kramer(1998)]{kramer1998directed}
Gerhard Kramer.
\newblock \emph{Directed information for channels with feedback}.
\newblock PhD thesis, Eidgenossiche Technische Hochschule Zurich, 1998.

\bibitem[Lillicrap et~al.(2015)Lillicrap, Hunt, Pritzel, Heess, Erez, Tassa,
  Silver, and Wierstra]{lillicrap2015continuous}
Timothy~P Lillicrap, Jonathan~J Hunt, Alexander Pritzel, Nicolas Heess, Tom
  Erez, Yuval Tassa, David Silver, and Daan Wierstra.
\newblock Continuous control with deep reinforcement learning.
\newblock \emph{arXiv preprint arXiv:1509.02971}, 2015.

\bibitem[Lopes et~al.(2012)Lopes, Lang, Toussaint, and
  Oudeyer]{lopes2012exploration}
Manuel Lopes, Tobias Lang, Marc Toussaint, and Pierre-Yves Oudeyer.
\newblock Exploration in model-based reinforcement learning by empirically
  estimating learning progress.
\newblock In \emph{Advances in Neural Information Processing Systems}, pages
  206--214, 2012.

\bibitem[Mohamed and Rezende(2015)]{mohamed2015variational}
Shakir Mohamed and Danilo~Jimenez Rezende.
\newblock Variational information maximisation for intrinsically motivated
  reinforcement learning.
\newblock In \emph{Advances in neural information processing systems}, pages
  2125--2133, 2015.

\bibitem[Mont{\'u}far et~al.(2016)Mont{\'u}far, Ghazi-Zahedi, and
  Ay]{montufar2016information}
Guido Mont{\'u}far, Keyan Ghazi-Zahedi, and Nihat Ay.
\newblock Information theoretically aided reinforcement learning for embodied
  agents.
\newblock \emph{arXiv preprint arXiv:1605.09735}, 2016.

\bibitem[Pathak et~al.(2017)Pathak, Agrawal, Efros, and
  Darrell]{pathak2017curiosity}
Deepak Pathak, Pulkit Agrawal, Alexei~A Efros, and Trevor Darrell.
\newblock Curiosity-driven exploration by self-supervised prediction.
\newblock \emph{arXiv preprint arXiv:1705.05363}, 2017.

\bibitem[Schmidhuber(1991)]{jurgen1991possibility}
J\H{u}rgen Schmidhuber.
\newblock A possibility for implementing curiosity and boredom in
  model-building neural controllers.
\newblock In \emph{From animals to animats: proceedings of the first
  international conference on simulation of adaptive behavior (SAB90)}, 1991.

\bibitem[Sutton and Barto(1998)]{sutton1998reinforcement}
Richard~S Sutton and Andrew~G Barto.
\newblock \emph{Reinforcement learning: An introduction}, volume~1.
\newblock MIT press Cambridge, 1998.

\bibitem[Tiomkin and Tishby(2017)]{tiomkin2017unified}
Stas Tiomkin and Naftali Tishby.
\newblock A unified bellman equation for causal information and value in markov
  decision processes.
\newblock \emph{arXiv preprint arXiv:1703.01585}, 2017.

\end{thebibliography}

\end{document}